# LMILAtt: A Deep Learning Model for Depression Detection from Social Media Users Enhanced by Multi-Instance Learning Based on Attention Mechanism

1st Yukun Yang
*School of Management*
*Beijing Institute of Technology*
Beijing, China
yangyukun2005@163.com

*Abstract*—Depression is a major global public health challenge and its early identification is crucial. Social media data provides a new perspective for depression detection, but existing methods face limitations such as insufficient accuracy, insufficient utilization of time series features, and high annotation costs. To this end, this study proposes the LMILAtt model, which innovatively integrates Long Short-Term Memory autoencoders and attention mechanisms: firstly, the temporal dynamic features of user tweets (such as depressive tendency evolution patterns) are extracted through unsupervised LSTM autoencoders. Secondly, the attention mechanism is used to dynamically weight key texts (such as early depression signals) and construct a multi-example learning architecture to improve the accuracy of user-level detection. Finally, the performance was verified on the WU3D dataset labeled by professional medicine. Experiments show that the model is significantly better than the baseline model in terms of accuracy, recall and F1 score. In addition, the weakly supervised learning strategy significantly reduces the cost of labeling and provides an efficient solution for large-scale social media depression screening.

*Keywords—text classification, weakly supervised learning, time series feature, multi-instance learning, attention mechanism*

I. INTRODUCTION

*A. Background*

Depression is a major global public health challenge, with typical symptoms including persistent depressed mood, sleep disturbances, distraction, and decreased interest in life, which can lead to somatic symptoms in severe cases [1]. According to the World Health Organization, between 2005 and 2015, the number of people with depression surged by 18.4% worldwide, and the number of confirmed cases exceeded the 332 million mark.[1] Also, according to the World Health Organization's 2019 data, suicide has become the second leading cause of death among people aged 15-29, and suicidal thoughts caused by depression are becoming increasingly common among adolescents.[2] Therefore, early identification of depressive symptoms is also critical to reducing the risk of suicide.

*B. Challenges*

Existing studies have made significant progress in using social media data for depression detection, but still face some limitations.

Firstly, some existing studies[2], [3] only focus on classifying a single piece of text, and cannot be directly applied to user-level depression detection; otherwise, it is easy to lead to misjudgment. After all, normal users sometimes express depression, and depressed users may also post tweets that look normal.

Secondly, some models [4], [5] were limited by their own structure and parameters, and only a small number of data samples were used. Therefore, trained models are prone to inaccuracy or insufficient generalization performance.

More importantly, some studies simply use a single-text classification model to identify the probability of depression in each text separately, and then average or maximize the output results. Although this approach is also conducive to improving accuracy, it cannot give attention to texts with different information densities, and ignores the features of texts in time series.

*C. Contributions*

To address the above limitations, this study proposes a new framework for social media-based detection of depression, the marginal contribution of which is summarized as follows.[3]

- A novel deep learning model for detecting depressed social media users is proposed. Through systematic comparison and verification, the score of the model on a large public dataset labeled according to professional medical criteria is significantly better than that of the baseline models, which achieve 96.95% accuracy, 97.30% precision, 94.60% recall, and 96.88% F1-score, so it has been proven to have high technological advancement and practical value.

- Our model innovatively put a multi-instance learning method by combining the attention mechanism and the time series analysis of the LSTM autoencoder to extract the temporal dynamic characteristics of user tweets. Through this method, the importance of each tweet is dynamically weighted using attention mechanisms, such as giving higher weight to tweets that signal early depression or key mood swings.

- The model adopts a weakly supervised learning strategy, which only requires binary labeling at the user level and does not require fine-grained annotation of each tweet. No labels are required at all during the text embedding and LSTM autoencoder feature extraction stages, and only user-level labels are used during final classifier training. This approach significantly reduces the cost of data annotation, avoiding the heavy lifting of manually annotating each tweet in traditional methods.

---

[1] https://www.who.int/en/news-room/fact-sheets/detail/depression
[2] https://www.who.int/news-room/fact-sheets/detail/suicide

[3] The code and data used in this work are available at the website: https://github.com/yangyukun2005/LMILAtt2

The rest of the article is organized as follows. In Section II, the related works in the field of depression detection are reviewed. Section III introduces principles of the model proposed in this study. Section IV details the experimental design and results. Finally, Section V summarizes the full text and proposes the next stage of our research.

## II. RELATED WORK

The research on social network depression detection based on artificial intelligence is mainly divided into two directions: artificial feature extraction and traditional machine learning model construction, and automatic feature extraction and deep neural network model construction. Notably, some deep learning-based detection methods introduce machine learning methods to further improve their performance. The work of each research direction is described separately below.

### A. Methods based on traditional machine learning

Depression detection methods based on traditional machine learning usually rely on expert knowledge to design feature engineering, including text emotional features, social behavior features, basic user information, etc., and then use classical machine learning algorithms to classify them. Its advantages are that the model is highly interpretable and requires relatively low computing resources, but its performance is highly dependent on the quality of feature engineering, and it is difficult to automatically mine deep patterns in the data.

De Chowdhury et al. [6] have pioneered a breakthrough in this field of research. By analyzing the behavioral features of Twitter users, they built a detailed feature engineering analysis process and a clear modeling methodology. Subsequently, Wang et al. [7] used sentiment analysis techniques to assess depressive tendencies in tweets with the help of artificially formulated vocabulary rules. Research shows that text-based features play a key role in online depression detection.

Deshpande et al. [3] proposed a natural language processing-based representational learning method for modeling Twitter text, and their approach allows classifiers to automatically capture potential features. Subsequently, Islam et al. [8] proposed a more advanced algorithm that uses Facebook data, uses LIWC to process text data, introduces KNN to classify text for depression, and can effectively learn the potential sparse representation of user features. Mustafa et al. [4] proposed for the first time a neural network model for detecting depressed users on social networking platforms.

### B. Methods based on deep learning

The deep learning-based depression detection method automatically learns the feature representations in the data through neural networks, avoiding the tedious process of manual feature engineering. This method can automatically extract multi-level feature representations from raw data, especially for processing high-dimensional, unstructured social media data. Deep learning methods excel in handling complex pattern recognition tasks, capturing nonlinear relationships and long-distance dependencies in data, and have made significant progress in the field of depression detection in recent years.

Yoon et al. [9] constructed a depression dataset containing 951 users' tweets and used LSTM to capture the features of each tweet, then averaged them and input them to the classifier. Subsequently, Ghosh et al. [10] proposed an attention-based model of BiLSTM-CNN to make the most of the information in each user's Tweet and give them reasonable weights. Mann et al. [11] both transformed the depression recognition task into a multi-instance learning question and used multimodal social media data for detection. They constructed a glossary of depression-related terms that conformed to the features of the Chinese language, extracted the features of depression-related terms, such as frequency and emotional tendency, and then input them into multiple machine learning models for weighted fusion, and finally input the fusion features into the LSTM.

In recent years, more advanced research based on deep learning has been proposed. Zhang et al. [12] proposed a knowledge-aware deep learning model called DKDD (Deep Knowledge-aware Depression Detection), which extracts clinically significant entities and their temporal distribution features from social media users' data through entity recognition, medical ontology alignment, and attention mechanisms. Yan et al. [13] propose a deep learning model called Depressive Emotion-Context Enhanced Network (DECEN), which combines a depressive emotion recognition module and a context-aware representation mechanism to improve the ability to detect depression from social media content and achieve a 93.05% precision. Kuang et al. [14] proposed an interpretable method called Multi-Scale Temporal Prototype Network (MSTPNet), which uses a pre-trained large language model to convert tweets into vectors, divide tweets into different time periods, calculate the intensity of symptoms on different time scales, and finally judge depressive status based on multi-scale symptom distribution.

## III. METHODOLOGY

First, let's define the research question: Suppose that in the dataset, there are $k$ social media users, which can be represented as $U = [u_1, u_2, ..., u_k]$, and each user posts more than 1 tweet. We suppose that the user $u_i$ publishes $m$ tweets, and arranges each text in chronological order from front to back, which can be expressed as $Text_i = [T_1, T_2, ..., T_m]$. We expect to detect whether the user $u_i$ is depressed.

In order to achieve the above goals, this study proposes a depression detection model of social media users using LSTM (long short-term memory) and a multi-instance learning method based on attention mechanisms. We name it LMILAtt (Long Short-Term Memory with Multi-Instance Learning by Attention). It consists of four modules: 1. Use the text embedding model to convert multiple tweets posted by social media users into vectors; 2. Use the LSTM autoencoder to extract time series features from text vectors; 3. Use the attention mechanism to realize the use of features of multiple texts with different weights and generate aggregate feature vectors; 4. Use aggregate feature vectors to train the classifier. Fig. 1 below shows the overall framework of this model.

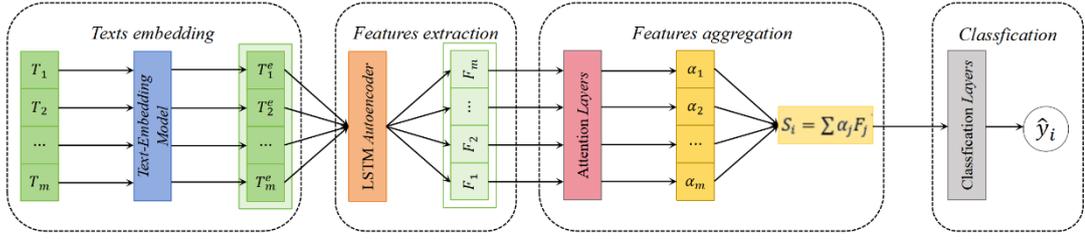

Fig. 1. Framework of our model

## A. Text preprocessing and embedding

First, we preprocess the text as follows: 1. Remove HTML tags (if any); 2. Remove special characters that are neither Chinese characters nor English letters nor punctuation; 3. Remove extra spaces in the text and spaces at the beginning and end; 4. Convert Chinese Traditional to Chinese Simplified, 5. Convert English uppercase letters to lowercase, and 6. Remove stop words.[4]

Then, we use the Qwen3-Embedding-0.6B large language model to embed each tweet posted by users at different times in the dataset as text vectors.[5] Taking the user $u_i$ as an example, if each text published by him/her is arranged in chronological order as $Text_i = [T_1, T_2, \ldots, T_m]$, (where $T_j$ represents the j-th text published by it, $j = 1, 2, \ldots, m$), then the vector set obtained after embedding can be expressed as $Text_i^e = [T_1^e, T_2^e, \ldots, T_m^e]$, (where $T_j^e$ represents the vector obtained by $T_j$, $j = 1, 2, \ldots, m$).

## B. Feature extraction

After embedding user text as a vector, to extract information from the user's timeline tweets, we utilize an unsupervised LSTM (also known as an LSTM autoencoder) to extract a set of vectors from each tweet. Among them, the autoencoder is a reconstructed neural network that learns the vectorized representation of each tweet in an unsupervised manner.

LSTM autoencoders require at least two LSTM layers. Taking the two LSTM layer models as an example, the first layer can be regarded as an encoder, and the second layer can be regarded as a decoder. First, the model inputs a vectorized tweet through the first LSTM layer to output a well-shaped vector. Then, take this vector as input so that through the 2nd LSTM layer, the output has the same shape as the input vectorized tweet; Finally, the model will be optimized to make the inputs and outputs as similar as possible.

Taking the input $Text_i^e = [T_1^e, T_2^e, \ldots, T_m^e]$ of the user $u_i$ as an example. After the above process, we get the tweet time-series features $TSF_i = [F_1, F_2, \ldots, F_m]$. where $T_j^e$ represents the input text vector, $F_j$ represents the output feature, $j = 1, 2, \ldots, m$.

## C. Aggregation of features

Previously, the time series features of the user $u_i$ have been extracted as $TSF_i = [F_1, F_2, \ldots, F_m]$. To make the most of the information from each tweet and distribute attention to different Tweets, we use attention mechanics. The formula for calculating the attention score for the j-th feature is shown in Equation(1).

$$\alpha_j = v^T \cdot tanh(WF_j + b) \quad (1)$$

where $v^T$ represents the attention vector, $W$ represents the weight matrix, $b$ represents the bias, and $tanh(\cdot)$ represents the hyperbolic tangent function.

After calculating the attention score for each feature, they are normalized. As shown in Equation(2).

$$\alpha_j = \frac{exp(\alpha_j)}{\sum_{k=1}^{m} exp(\alpha_k)} \quad (2)$$

Then, the weighted aggregation of each feature is performed to obtain the aggregated feature vector:

$$S_i = \sum_{j=1}^{m} \alpha_j \cdot F_j \quad (3)$$

## D. User classification

After getting the aggregated feature vector $S_i$ of the user $u_i$, it is entered into the fully connected layer:

$$z_i = W_f S_i + b_f \quad (4)$$

where $W_f$ represents the weight matrix, and $b_f$ represents the bias.

After that, it is entered into the sigmoid activation function so that it is mapped between 0 and 1:

$$\hat{y}_i = sigmoid(z_i) = \frac{1}{1 + e^{-z_i}} \quad (5)$$

where the obtained $\hat{y}_i$ is the probability of the user's depression.

To train the classifier, we define the loss function as Equation(6) and use the Adam optimizer to adjust the model parameters to minimize the value of the loss function.

$$loss = -\frac{1}{N}[y_i \log(\hat{y}_i) + (1 - y_i)\log(1 - \hat{y}_i)] \quad (6)$$

where $y_i$ represents the real label of the user $u_i$, and $N$ represents the number of users included in the training set.

## IV. EXPERIMENT AND EVALUATION

### A. Software and hardware environment

All of the following experiments were performed in the environment shown in Table I.

### B. Dataset description

This study uses a large public dataset called WU3D (Weibo User Depression Detection Dataset)[6][15]. The dataset is derived from Sina Weibo, China's largest social media platform, and contains tags about whether users are depressed. All annotated data are based on the Diagnostic and Statistical

---

[4] https://gitcode.com/open-source-toolkit/595be
[5] https://modelscope.cn/models/Qwen/Qwen3-Embedding-0.6B
[6] https://github.com/aidenwang9867/Weibo-User-Depession-Detection-Dataset

Manual of Mental Disorders, Fifth Edition (DSM-5) and have been reviewed twice by psychologists and psychiatrists. To ensure anonymity, all username information has been replaced. The overall description of this dataset is shown in Table II. In this experiment, 5000 normal users and 5000 depressed users were selected as subsets, and randomly divided into training set, validation set, and test set according to the ratio of 6:2:2.

TABLE I. HARDWARE AND SOFTWARE SETTINGS

| Type | Items | Settings |
|---|---|---|
| Hardware settings | Operating system | Windows 11 (64-bit) |
|  | Graphics | NVIDIA GeForce RTX 3060 |
| Software settings | PyCharm | v3.1.1 |
|  | Anaconda | v3.12.0 |
|  | torch | v2.8.0+cu129 |
|  | torchvision | v0.23.0+cu129 |
|  | modelscope | v1.29.1 |
|  | scikit-learn | v1.7.1 |
|  | matplotlib | v3.10.5 |
|  | numpy | v1.26.4 |
|  | seaborn | v0.13.2 |
|  | pandas | 2.2.3 |

TABLE II. DESCRIPTION OF WU3D

| Labels | Number of users | Total tweets | Average number of tweets per user | Average length of each tweet |
|---|---|---|---|---|
| Normal | 21872.00 | 1534987.00 | 70.33 | 52.70 |
| Depression | 9865.00 | 359871.00 | 36.48 | 100.25 |
| Total | 31737.00 | 1894858.00 | 59.71 | 61.73 |

C. Model effect evaluation indicators

In the field of binary classification, the evaluation of model effect mainly adopts four indicators: accuracy, precision, recall, and F1-score. They are calculated as follows:

$$accuracy = \frac{TP + TN}{TP + TN + FN + FP} \quad (7)$$

$$precision = \frac{TP}{TP + FP} \quad (8)$$

$$recall = \frac{TP}{TP + FN} \quad (9)$$

$$F1\_score = \frac{2 \times precision \times recall}{precision + recal} \quad (10)$$

The above TP (True Positive) represents the number of depressed users who are correctly predicted, TN (True Negative) represents the number of normal users who are correctly predicted, FP (False Positive) represents the number of normal users who are incorrectly predicted, and FN (False Negative) represents the number of depressed users who are incorrectly predicted.

D. Baseline Models

To verify the performance of the proposed deep learning method, we set the following four baseline models:

*1) Naive Bayes.* It is a simple probability classification algorithm based on Bayes' theorem, which assumes that features are independent of each other and is often used as a baseline model in similar tasks [16], [17], [18].

*2) Random Decision Forest.* It is a machine learning algorithm that integrates multiple classifiers through ensemble learning and is widely used in similar tasks [4], [19], [20].

*3) Multi-Channel-CNNAtt-LSTMAtt* [18]. It first segments the text and then uses Word2Vec to output the word vector. Next, the CNN-Attention layer is used to feature the text. Input the text word vector matrix, and then use multiple convolutional checks of different sizes to convolve the matrix, and the convolutional results are pooled to the maximum to obtain several new eigenvectors. Then, each feature vector is used as the input of multi-channel attention, and the matching score of each feature vector is calculated, and then the percentage of each feature vector score is calculated separately. After obtaining the percentage corresponding to each eigenvector, the weighted sum of all eigenvectors is performed to obtain the overall text vector. Finally, the overall text vector is input into the LSTM-Attention layer to obtain the final result, and it is input into the activation function to realize the binary classification of text.

*4) Multi-Scale Temporal Prototype Network* [14]. It is an innovative, explainable model for depression detection in social media. This method innovatively detects and interprets depressive symptoms and their duration, meeting clinical diagnostic criteria. MSTPNet first uses the pre-trained large language model BERT-base-Chinese to convert social media tweets into semantic vectors, and then divides the tweets into different time periods according to semantic similarity and time intervals through the time segmentation layer. Then, the multi-scale temporal prototype layer compares the learned prototypes of depressive symptoms with tweets in each time period to calculate the intensity of the presence of symptoms on different time scales. Finally, the classification layer judges the depressive state based on these multi-scale symptom distributions.

E. Experimental results

*1) Scores of proposed model*: Fig. 2 shows the training process of this model. When the training period is less than 11, the loss value of the model on the validation set decreases slightly in fluctuations, but when the training period is larger, the loss value rises rapidly. Therefore, the optimal training period for the model is around 11. Fig. 3 shows the ROC curve of the model. Fig. 4 shows the confusion matrix.

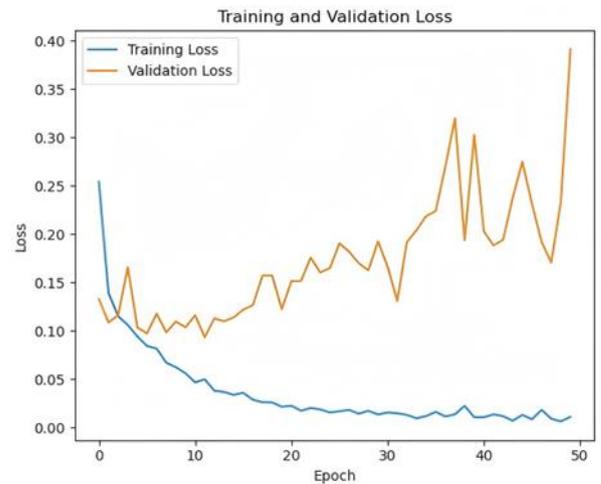

Fig. 2. Loss value under different training epochs

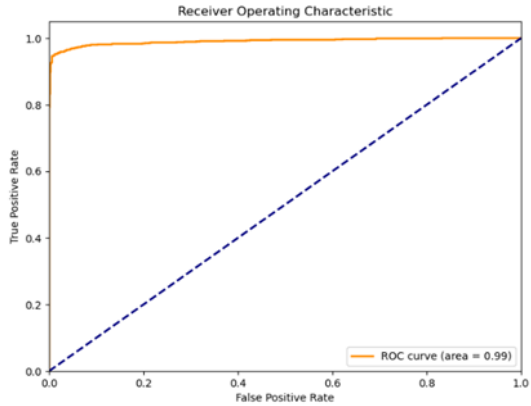

Fig. 3. ROC curve

Fig. 4. Confusion Matrix

*2) Comparison of different models*: Table III shows the performance comparison between two classical machine learning models, two more advanced deep learning models, and the proposed method.

TABLE III. SCORES OF DIFFERENT METHODS

| Models | Accuracy | Precision | Recall | F1_Score |
|---|---|---|---|---|
| NB | 0.8632 | 0.8502 | 0.8807 | 0.8652 |
| RDF | 0.8745 | 0.8692 | 0.8804 | 0.8748 |
| MCCL | 0.9166 | 0.9016 | 0.9347 | 0.9179 |
| MSTPNet | **0.9710** | 0.9570 | 0.7660 | 0.8510 |
| Our Model | 0.9695 | **0.9927** | **0.9460** | **0.9687** |

## V. CONCLUSION

In this study, the LMILAtt model, a deep learning framework designed for the detection of depression in social media users, is proposed and validated. Through the LSTM autoencoder, the model can capture the temporal dynamic characteristics of user tweet sequences, revealing the underlying evolution patterns of depressive tendencies. The attention mechanism gives the model the ability to dynamically evaluate the information density of different tweets, giving higher weight to key signals, and effectively avoiding the problem of misjudgment caused by single-text classification.

Comparative experiments on the large labeled dataset WU3D show that LMILAtt is significantly better than multiple baseline models in terms of accuracy, recall, precision and F1 score for depression detection at the user level. In addition, the weakly supervised learning strategy adopted by the model only requires user-level binary labels, which greatly reduces the cost of data annotation and significantly improves the scalability and practical application efficiency of the model.

Future research directions may be developed from the following aspects: 1. Integrate multimodal data (such as social images or videos) to capture more comprehensive depression signals by expanding cross-modal attention mechanisms. 2. Enhance model interpretability and draw on MSTPNet's clinical alignment methods (such as symptom prototype analysis) to make the decision-making process transparent and comply with diagnostic criteria.